%% file: main.tex
\documentclass[lettersize,journal]{IEEEtran}
\usepackage{amsmath,amsfonts}
\usepackage{algorithmic}
\usepackage{algorithm}
\usepackage{array}
\usepackage[caption=false,font=normalsize,labelfont=sf,textfont=sf]{subfig}
\usepackage{textcomp}
\usepackage{stfloats}
\usepackage{url}
\usepackage{verbatim}
\usepackage{graphicx}
\usepackage{cite}
\usepackage{booktabs}           % \toprule \midrule \bottomrule
\usepackage{multirow}           % \multirow
\usepackage[table]{xcolor}      % \cellcolor + 颜色表格；也提供 \textcolor
\usepackage{makecell}           % \thead
\usepackage{capt-of}            % \captionof{table} / \captionof{figure}  (不要用 caption 包)

\begin{document}

\title{WaterWave: Bridging Underwater Image Enhancement \\ into Video Streams via Wavelet-based Temporal Consistency Field}

\author{Qi~Zhu,
        Jingyi~Zhang,
        Naishan~Zheng,
        Wei~Yu,
        Jinghao~Zhang,
        Deyi~Ji,
        and~Feng~Zhao% <-this % stops a space
\thanks{Qi Zhu, Naishan Zheng, Wei Yu, Jinghao Zhang, Deyi Ji, and Feng Zhao are with the University of Science and Technology of China (e-mail: zqcrafts@mail.ustc.edu.cn).}%
\thanks{Jingyi Zhang is with Hefei University of Technology.}%
\thanks{Corresponding author: Feng Zhao (e-mail: fzhao956@ustc.edu.cn).}}

% The paper headers
% \markboth{Journal of \LaTeX\ Class Files,~Vol.~14, No.~8, August~2021}%
% {Shell \MakeLowercase{\textit{et al.}}: A Sample Article Using IEEEtran.cls for IEEE Journals}

% \IEEEpubid{0000--0000/00\$00.00~\copyright~2021 IEEE}
% Remember, if you use this you must call \IEEEpubidadjcol in the second
% column for its text to clear the IEEEpubid mark.

\maketitle

\input{sec/0_abstract}

\vspace{2mm}
\vspace{-2mm}
% \section{Introduction}

\input{sec/1_intro}

\input{sec/2_related}
\input{sec/3_motivation}

\input{sec/4_method}
\input{sec/5_exer}

\bibliographystyle{IEEEtran}
\bibliography{main}
\vspace{-30pt}
% \input{sec/bio}

% \newpage

% \section{Biography Section}
% If you have an EPS/PDF photo (graphicx package needed), extra braces are
%  needed around the contents of the optional argument to biography to prevent
%  the LaTeX parser from getting confused when it sees the complicated
%  $\backslash${\tt{includegraphics}} command within an optional argument. (You can create
%  your own custom macro containing the $\backslash${\tt{includegraphics}} command to make things
%  simpler here.)
 
% \vspace{11pt}

% \bf{If you include a photo:}\vspace{-33pt}
% \begin{IEEEbiography}[{\includegraphics[width=1in,height=1.25in,clip,keepaspectratio]{fig1}}]{Michael Shell}
% Use $\backslash${\tt{begin\{IEEEbiography\}}} and then for the 1st argument use $\backslash${\tt{includegraphics}} to declare and link the author photo.
% Use the author name as the 3rd argument followed by the biography text.
% \end{IEEEbiography}

% \vspace{11pt}

% \bf{If you will not include a photo:}\vspace{-33pt}
% \begin{IEEEbiographynophoto}{John Doe}
% Use $\backslash${\tt{begin\{IEEEbiographynophoto\}}} and the author name as the argument followed by the biography text.
% \end{IEEEbiographynophoto}

\vfill

\end{document}

%% file: sec/0_abstract.tex
\begin{abstract}
Underwater video pairs are fairly difficult to obtain due to the complex underwater imaging. In this case, most existing video underwater enhancement methods are performed by directly applying the single-image enhancement model frame by frame, but a natural issue is lacking temporal consistency. To relieve the problem, we rethink the temporal manifold inherent in natural videos and observe a temporal consistency prior in dynamic scenes from the local temporal frequency perspective.
Building upon the specific prior and no paired-data condition, we propose an implicit representation manner for enhanced video signals, which is conducted in the wavelet-based temporal consistency field, WaterWave.
Specifically, under the constraints of the prior, we progressively filter and attenuate the inconsistent components while preserving motion details and scenes, achieving a natural-flowing video. Furthermore, to represent temporal frequency bands more accurately, an underwater flow correction module is designed to rectify estimated flows considering the transmission in underwater scenes. Extensive experiments demonstrate that WaterWave significantly enhances the quality of videos generated using single-image underwater enhancements.
Additionally, our method demonstrates high potential in downstream underwater tracking tasks, such as UOSTrack and MAT, outperforming the original video by a large margin, i.e., 19.7\% and 9.7\% on precise respectively. The code is publicly available to facilitate further research at \url{https://github.com/zqcrafts/WaterWave}.

\end{abstract}
\vspace{-3mm}

%% file: sec/1_intro.tex
\begin{figure*}[t!]
\centering
\hspace{2mm} \includegraphics[width=\textwidth]{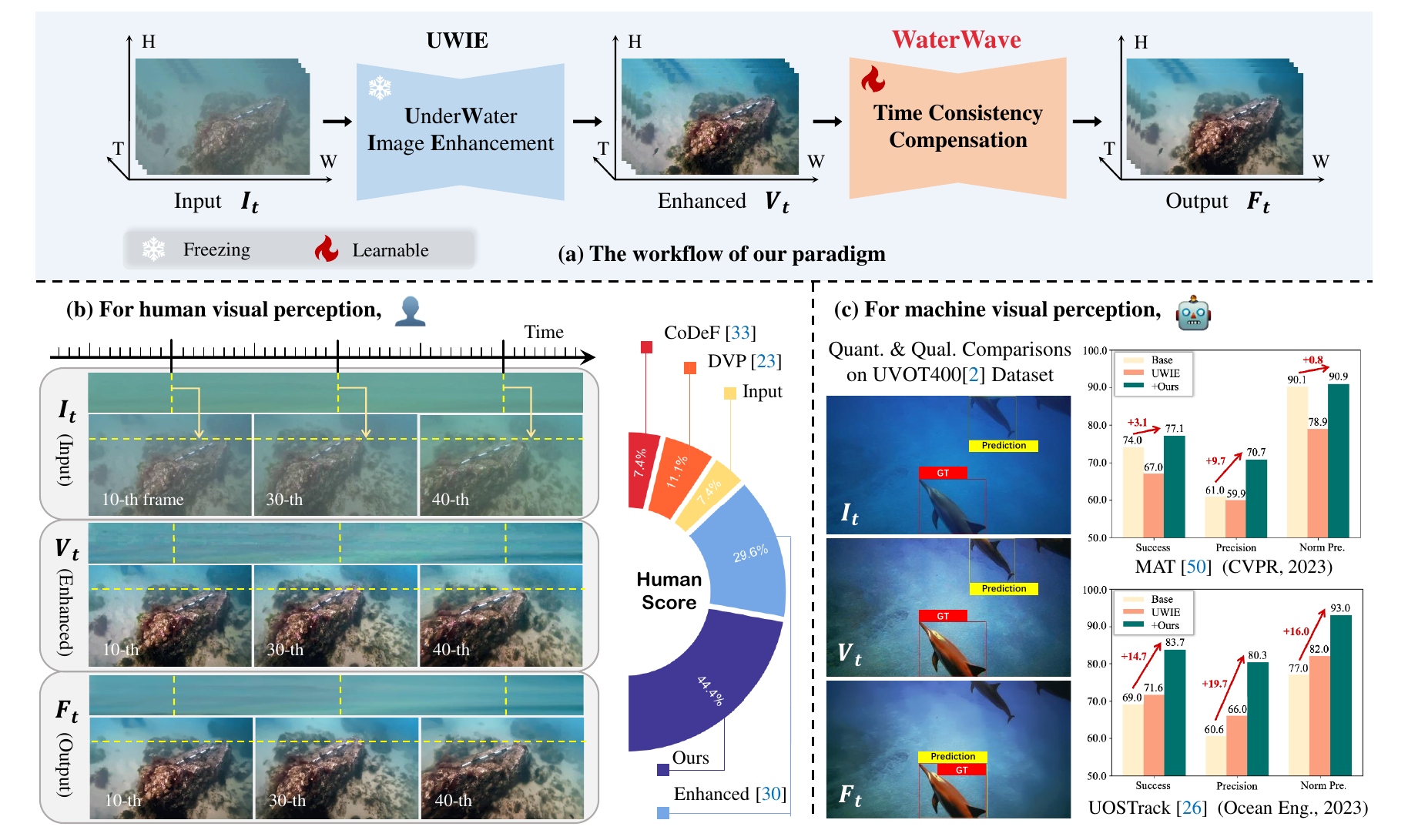}
\vspace{-3mm}
\caption{
(a) The workflow of our paradigm. (b) For human visual perception, our method recovers the temporal consistency from UWIE enhanced results and harvest the favor of the majority.
(c) For machine visual perception, the proposed temporal consistency compensation significantly surpasses the baseline while UWIE may even negatively affects tracking performance due to the absence of temporal consistency.}
\vspace{-3mm}
\label{fig1}
\end{figure*}

\section{Introduction}

Underwater video enhancement is crucial for improving the visibility of submerged imagery, facilitating advanced marine research, and enabling more effective autonomous underwater navigation and monitoring systems. However, obtaining underwater video pairs can be significantly difficult since complex underwater imaging caused by wavelength-dependent light absorption and scattering, which limits the advancement of underwater video enhancement in the data-driven intelligence era.
Fortunately, in recent years, the Under-Water Image Enhancement (UWIE)~\cite{Li_prior_2020,Retinex_2019,Fast_2020,bilateral_2013,illuminated_2017,Li_Guo_2019} has made substantial progress by some mature vision framework and widely-used benchmarks.
% For example, Li~\textit{et. al}~\cite{Li_Guo_2019} construct a large-scale real-world underwater image enhancement benchmark, which obtains diverse color ranges and degrees of contrast decrease, greatly promoting the development of the underwater single-image enhancement. 

This leads to our main claim: \textit{``Can these image enhancement methods be leveraged gracefully to obtain high-quality enhanced videos?"}.
A naïve option, which also is employed by most existing underwater video enhancement methods~\cite{System_2020, fusion_2012,stereo_2015,fusiondehazing_2016, Li_prior_2020,Retinex_2019,Fast_2020,bilateral_2013,illuminated_2017}, would be directly applying existing single image enhancement algorithms frame by frame and then connected into video, where temporal inconsistent artifacts may occur as shown in Figure~\ref{fig1}(b).

% To relieve this problem, prolific algorithms~\cite{Lai_2018,ouyang2023codef,Lei_2020,Mrak_2023, Zhao_2023,Shen_Qi_2022,Unger_2019,Abbeel_2022,Pfister_2015,Eilertsen_2019} delve into compensating the blind temporal consistency following the UWIE network, as shown in Figure~\ref{fig1}(a).
% Nevertheless, these methods formulate temporal consistency independently and lacks of considering the underwater condition, like inaccurate underwater optical flow estimation and severe blocking noises in the depth water, which brings challenges for temporal modeling in underwater scenarios. 
% % insight：：：   prior  --》  把不连续噪声移除 --》 inr内插  ---》连续视频
% To address the limitation, we first observe the temporal inconsistency prior based on the space-temporal wavelet decomposition. Then, motivated by the powerful video signal fitting and interpolation ability of the implicit neural network (INR)~\cite{Brualla_2021,Chen_Shrivastava,Kwan_2023,Austin,Shabtay_2022,Chen_2021,Aiyetigbo2023,Zhang_2023,ouyang2023codef}, we re-fit the target signal under the temporal consistency prior, resulting the temporal continuous and enhance-maintained video, as shown in Table~\ref{tab1}.  

To relieve this problem, prolific algorithms~\cite{Lai_2018,ouyang2023codef,Lei_2020,Mrak_2023, Zhao_2023,Shen_Qi_2022,Unger_2019,Abbeel_2022,Pfister_2015,Eilertsen_2019} delve into compensating the blind temporal consistency. 
Nevertheless, these methods formulate temporal consistency independently and lack considering underwater-aware knowledge, like inaccurate underwater optical flow estimation and severe blocking noises in the depth water, which brings great challenges.
% insight：：：   prior  --》  把不连续噪声移除 --》 inr内插  ---》连续视频
To address the limitation, motivated by the powerful video signal fitting and interpolation ability of the implicit neural representation (INR)~\cite{Brualla_2021,Chen_Shrivastava,Kwan_2023,Austin,Shabtay_2022,Chen_2021,Aiyetigbo2023,Zhang_2023,ouyang2023codef}, we introduce a novel implicit re-representation manner, which learns to re-represent the original enhanced video from zero (\textit{i.e.} without any content) and gains underwater contents and temporal consistency during the representation process.
Unlike standard INRs, which only pursue the high fidelity of the original signal, we decently design a temporal consistency prior and require the prior to be met during the fitting process, achieving the desired video.

Specifically, based on the wavelet theory, we prove the temporal consistency prior that most temporal inconsistency may exist with spatial low-frequency regions. Considering this prior and underwater scenes, we design a temporal consistency underwater field, WaterWave, which is mainly driven by Transmission-guided Flow Rectification (TFR) and Video Consistency-aware Wavelet (VC-Wave) block.
For the TFR module, since flow estimation trained in open-air is not so effective in underwater, based on atmospheric scattering modeling, the transmission map is leveraged to rectify the original estimated flow for adapting underwater scenarios. Then, utilizing the temporal consistency prior, we design the VC-Wave block to decouple the video signals into the temporal inconsistency components and the basic contents. In the training stage, the inconsistency components are regularized, while the basic ones are still learned for high-fidelity. 
Leveraging the above delicate designs, the WaterWave encourages high-quality videos as shown in Figure~\ref{fig1}(b)(c).

% In conclusion, our contributions can be summarized as follows: 
% 1) We propose the first learning-based underwater video enhancement framework and pave a novel and potential path for the underwater video enhancement. 
% 2)  Our method complements the temporal consistency from results by existing underwater image enhancements.
% 3) We arouse the potential of UWIE methods in down-stream task, where nearly 10 and 19 improvement in precise on the MAT~\cite{MAT} and UOSTrack~\cite{UOSTrack} respectively.

% \vspace{-1mm}
In conclusion, our contributions can be summarized as follows: 
% \vspace{-1mm}
\begin{itemize}
    \item \noindent We propose the first learning-based underwater video enhancement framework and pave a novel and potential path for underwater video enhancement. 
    % \vspace{-1mm}
    \item \noindent Our method complements the temporal consistency in existing underwater image enhancement methods and achieves high-fidelity and fluency video.
    % \vspace{-1mm}
    \item \noindent We arouse the potential of UWIE methods in downstream tasks, where nearly 10\% and 19\% improvement in precision on the MAT~\cite{MAT} and UOSTrack~\cite{UOSTrack} respectively.
\end{itemize}

%% file: sec/2_related.tex
\section{Related Work}

\subsection{Underwater Video Enhancement}

With the ocean development of underwater video, underwater video has higher acquisition and application prospects than underwater images. 
Because of the optical properties, underwater video has some similar problems to underwater images, such as color bias, image blur, low contrast, uneven illumination, etc. Most of the existing underwater video enhancement methods are extensions of single image enhancement algorithms~\cite{Li_prior_2020,Retinex_2019,Fast_2020,bilateral_2013,illuminated_2017}. 
For example, UMCNN~\cite{Li_prior_2020} adopts a lightweight modular structure with the depth of only 10 layer, which is easy to train for use in frame-by-frame enhancement of underwater video. Tang et al.~\cite{Retinex_2019} optimize the computing strategy by fast filtering for application to underwater video.
However, simply enhancing video frames and connecting them will interrupt the correlations between adjacent frames.
To relieve this problem, some method use the timing characteristics combined with the timing relationship between frames~\cite{System_2020, fusion_2012,stereo_2015,fusiondehazing_2016}. Ancuti et al.~\cite{fusion_2012} use the time-bilateral filtering strategy and Qing et al.~\cite{fusiondehazing_2016} estimate the transmission image and atmospheric light value by temporal information. Li et al.~\cite{stereo_2015} perform video dehazing and stereo reconstruction simultaneously to deal with underwater video and obtain an excellent defogging effect. 
These traditional and CNN-based algorithms often sacrifice color consistency of the entire video or enhancement quality. Hence, how to strike a balance between video quality and continuity is an urgent problem to be solved.

\vspace{-5mm}
\subsection{Temporal Consistency Learning}

With the development of deep learning, significant progress has been made in image and video restoration~\cite{Zhu_2023_ICCV,Ren_2024_CVPR,yu2022source,li2024fouriermamba,zheng2023learning,yu2023learning,zheng2022enhancement,zhu2023learning,zhu2022dast,zhu2023fouridown,yu2024empowering,li2023frequency}.
Temporal consistency is a crucial indicator for evaluating high-quality videos. Recently, numerous video enhancement approaches~\cite{Zhao_2023,Shen_Qi_2022,Unger_2019,Abbeel_2022} achieve decent performance by the effective temporal modeling framework, which can be mainly divided into sliding-window and recurrent propagation manners.
However, these methods become difficult to be employed when lacking of paired video datasets. 
In this case, some researches~\cite{Pfister_2015,Lai_2018,Mrak_2023,Lei_2020,Eilertsen_2019} delve into blind video temporal consistency task that extends image processing techniques to videos. For instance, Bonneel~\textit{et al.}~\cite{Pfister_2015} propose a gradient domain technique to infer the temporal regularity as a temporal consistency guide to stabilize the processed sequence. With the development of deep learning, Lai~\textit{et al.}~\cite{Lai_2018} present an flow-based solution to remove temporal flickering via learning a deep ConvLSTM network. 
Moreover, DVP~\cite{Lei_2020} observes that temporal consistency can be achieved by training a video convolutional network and obtains good performances on blind video temporal consistency.
Nevertheless, these methods may be limited by the trade-off between temporal consistency and fidelity due to the inappropriate objective function.

\vspace{-4.3mm}

% \begin{figure*}[t]
% \centering
% \includegraphics[width=\textwidth]{fig/framework.pdf}
% \caption{Overview of WaterWave. 
% In the inference stage, all coordinates of video are given and the target video is generated. Essentially, an implicit neural network (INR) are performed, which consists of position encoding and MLP. 
% In the training stage, the network learns to fit the video signal meeting the temporal consistency prior, where given all coordinates 
% of adjacent frames to model video by the TFR module and the VC-Wave block. In this process, the temporal inconsistency is gradually regularized for fitting the high-quality and consistent video signal. 
% }
% \vspace{-2mm}
% \label{fig2}
% \end{figure*}

\subsection{Implicit Neural Representation}

Implicit neural representation (INR) is a novel way to fitting complicated natural signals such as images and videos~\cite{Brualla_2021,Chen_Shrivastava,Kwan_2023,Austin,Shabtay_2022,Chen_2021,Aiyetigbo2023,Zhang_2023,ouyang2023codef}. Given the pixel coordinate in a video, INR can map the coordinate to RGB value of the pixel.
Taking advantages of this property, some image and video restoration works use INR to extend the capabilities that are absent in the original network, such as arbitrarily resolutions~\cite{Austin,Shabtay_2022,Chen_2021,Aiyetigbo2023}.
Furthermore, different from convolution-based and transformer-based network, INR network learns targets from content-agnostic inputs~\textit{i.e.} coordinates.
Inspired by this, NeRCo~\cite{Zhang_2023} utilizes the neural representation to learn a uniform degradation level before the enhancement procedure. More recently, CoDeF~\cite{ouyang2023codef} compresses a clip of video into a canonical image by video INR and decode the image by the deformation field, which contains learned continuous motion information.
Compared with above methods, we are the first to leverage its controllable spatial-temporal fitting capability delicately to achieve blind underwater video temporal consistency.

%% file: sec/3_motivation.tex
\section{Motivation}
\subsection{Temporal Consistency Formulation}

In this section, we formulate the temporal consistency in videos and then analyze the relationship between temporal continuity and spatial frequency, which inspires the design of the proposed method.

First, we perform the 1-D wavelet transform for $t$ to decompose the video signal $F(t)$ into two bands as follow
% \begin{equation}
% F(t) =\sum_{k=-\infty}^{\infty} a(k) \phi(t-k)+
% \sum_{j=0}^{\infty} \sum_{k=-\infty}^{\infty} b(j, k) 2^{j / 2} \psi\left(2^j t-k\right),
% \end{equation}
{\small
\begin{equation}
F(t)=\sum_{k=-\infty}^{\infty} a(k)\,\phi(t-k)+
\sum_{j=0}^{\infty}\sum_{k=-\infty}^{\infty} b(j,k)\,2^{j/2}\,\psi(2^{j}t-k),
\end{equation}
}
where the former term is low-frequency part and the latter term is high-frequency part. To recover the temporal continuity, the low-pass filtering in the temporal domain~\cite{Paris_2008} is used to suppress high-frequency components, which can be expressed as
\begin{equation}
min H(j,k) = min \sum_{j=0}^{\infty} \sum_{k=-\infty}^{\infty} b(j, k) 2^{j / 2} \psi\left(2^j t-k\right).
\end{equation}
% However, regularizing uniformly to all temporal high-frequency bands may damage original motion details. Widely-used methods add spatial terms, as shown in Table~\ref{tab1}. where the object function can be written as
However, regularizing uniformly to all temporal high-frequency bands may damage original motion details. Widely-used methods add spatial terms, as shown in Table. where the object function can be written as
\begin{equation}
min \mathbb{L} = min L_2\left(S_V-S_F\right) + w \cdot L_2\left(H(j,k)\right),
\end{equation}
where $L_2$ is Euclidean norm loss, $S_V$ and $S_F$ are spatial characters of input video $V$ and output video $F$ respectively, and $w$ is a hyper-parameter. 
\subsection{Temporal Inconsistency Prior}
To simplify the proof, we use Haar wavelets as orthogonal basis, where wavelet function $\psi$ is
\begin{equation}
\psi(2^j t-k)= \begin{cases}1, & \text { for } 0 \leq t<1 / 2 \\ -1, & \text { for } 1 / 2 \leq t<1 . \\ 0, & \text { otherwise }\end{cases}
\end{equation}
% \begin{equation}
% \phi(x)= \begin{cases}1, & \text { for } 0 \leq x<1 \\ 0, & \text { otherwise }\end{cases}
% \end{equation}
And the wavelet coefficient $b(j, k)$ can be computed by
\begin{equation}
b(j, k)=2^{j / 2} \int_{-\infty}^{\infty} F(t) \psi\left(2^j t-k\right) d t.
\end{equation}
To ensure coherent illumination and motion effects, we consider two adjacent frames as $F=\{I_t(x),Warp(I_{t-1}(x)\}$ and choose spatial gradient as $C_s$. Considering $k \in [0,2^j-1)$, wavelet is supported when $k=0, j=0$. 
Combining Equation (2), (4) and (5), we use the Euler-Lagrange formula to minimize Equation (3) when $\frac{\partial \mathbb{L}}{\partial F}-\frac{d}{d x}\left(\frac{\partial \mathbb{L}}{\partial \nabla F}\right)=0$ and obtain that $I_t$ must satisfy following condition as:
\begin{equation}
-\Delta F_t+ w F_t=-\Delta V_t + w \cdot \textit{Warp}\left(F_{t-1}\right),
\end{equation}
where $\Delta$ denotes Laplace operator.
Then we implement Fourier transform $\mathcal{F}(.)$ on both sides of Equation (6) and take $\Delta \mathcal{F}(.)= -4 \pi^2 u^2 \mathcal{F}(\cdot)$ into the equation, set spatial frequency $v = 2 \pi u$ and $w = 1$, finally, the temporal consistent video can be calculated by:
\begin{equation}
\mathcal{F}\left(F_t\right)=
\frac{v^2}{v^2 +1}\mathcal{F}\left(V_t\right) +
\frac{1}{v^2+ 1}\mathcal{F}\left(\textit{Warp}\left(F_{t-1}\right)\right),
\end{equation}
where the former term implies scene fidelity and the latter term implies temporal consistency. The detailed derivation is referred in the supplementary. From Equation (7), it can be seen that as the spatial frequency $v$ decreases, temporal continuity should be improved while scene fidelity are opposite.
Motivated by this prior, we assume that the temporal inconsistency components may mainly exist with spatial low-frequency but temporal high-frequency regions.

%% file: sec/4_method.tex
\section{Methodology: WaterWave}

In this section, we elaborate on details of the proposed 
blind temporal consistency learning paradigm. First, as shown in Figure~\ref{fig1}(a), an input underwater video $I_t$ is enhanced frame by frame via the pretrained underwater image enhancement (UWIE) network, obtaining a well-enhanced but temporal inconsistent video, denoted as $E_t$. 
% Then, the proposed WaterWave is employed to process this inconsistent video. 

% As illustrated in Figure~\ref{fig2}, WaterWave consists of four components including position encoding, implicit neural representation, motion alignment and video continuity learning. 
% During the training stage, considering general temporal modeling, the coordinates of adjacent frames, $(x, y, t)$ and $(x, y, t+1)$, are served as inputs. In order to extract more precise temporal frequency, which is basic to the temporal consistency prior, the flow estimated between $I_t$ and $I_{t+1}$ are refined by the Transmission-guided 
% Flow Rectification (TFR) module.
% After motion warping, $(x, y, t)$ and $(x+\Delta x^r, y +\Delta y^r, t+1)$ are sent into the 3D multi-resolution hash table to obtain the position encoding. 
% Subsequently, the position encoding is fed into multi-layer perception (MLP) network to fit the desired temporal consistent video signal $F_t$. In the learning process, motivated by the temporal consistency prior, the VC-Wave is designed to decoupled into consistency and inconsistency components from video, where the former is maintained and the latter is annealed.
% During the inference stage, note that the alignment and the learning part are dropped, just input all coordinates of target video and a high-fidelity and fluency video is output.
% The details of the main modules are explained next.

\begin{figure*}[t]
\centering
\includegraphics[width=\textwidth]{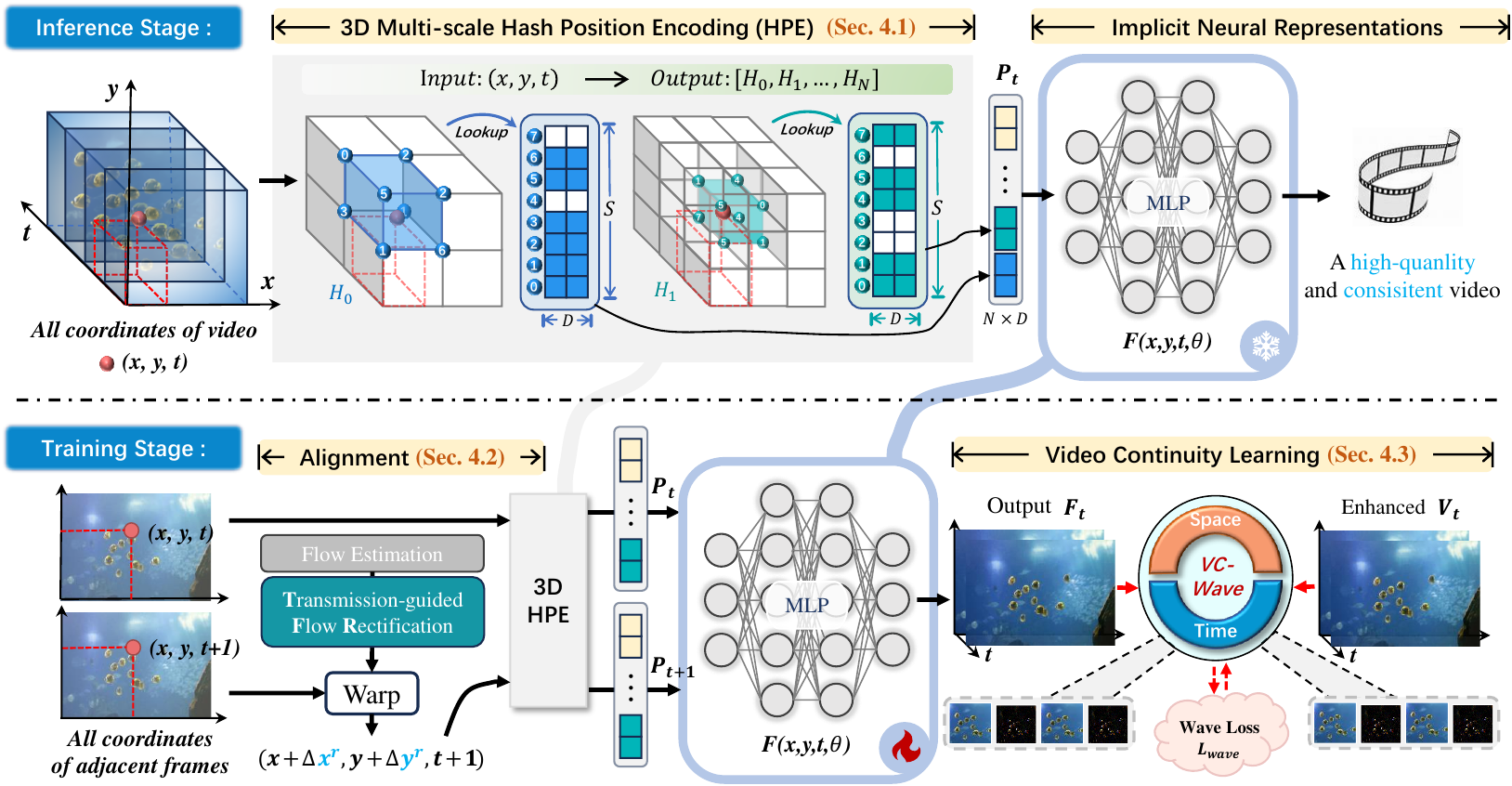}
\caption{Overview of WaterWave. 
In the inference stage, all coordinates of video are given and the target video is generated. Essentially, an implicit neural network (INR) are performed, which consists of position encoding and MLP. 
In the training stage, the network learns to fit the video signal meeting the temporal consistency prior, where given all coordinates 
of adjacent frames to model video by the TFR module and the VC-Wave block. In this process, the temporal inconsistency is gradually regularized for fitting the high-quality and consistent video signal. 
}
\vspace{-2mm}
\label{fig2}
\end{figure*}
\vspace{-4mm}
\subsection{3D Multi-scale Hash Position Encoding}
To more accurately represent the complex motion details in underwater, such as the undulating movements of fish and coral, we utilize a multi-scale hash positional encoding technique~\cite{Mller2022InstantNG}, which efficiently generates a latent positional code for each coordinate within a scene.
Specifically, as depicted in Figure~\ref{fig2}, within a 3D Cartesian coordinate system, a voxel in the video can be specified by its 3D coordinates \((x, y, t)\). We determine the neighboring voxels across multiple resolution levels, denoted as \([H_0, H_1, \ldots, H_N]\), where \(H_N\) represents the highest resolution. Each voxel's corner points are indexed by hashing their integer-based coordinates. For every corner index, we retrieve its associated multi-dimensional feature vector from the hash tables, where hash feature dimension is $D$ and hash table size is $S$. These vectors are then linearly interpolated based on the relative position of \(x\) within its respective resolution level. Finally, the sub-positional embedding from all levels are concatenated as output embedding, \(\mathbf{P} \in \mathbb{R}^{N \times F}\), providing a rich representation of the spatial-temporal information for subsequent processing.

Moreover, a high-quality underwater video usually comprises a blend of distinctive high and low frequency regions, for instance, the deep water exhibits strong low-frequencies whereas corals and fish present numerous high-frequencies. Inspired by this mind, we employ a hash encoding annealing strategy~\cite{Brualla_2021} that enables the implicit neural representation to differentially adapt to low and high frequencies throughout the training process, which is performed by a controllable parameter, $\alpha(k) = \frac{Nk}{s}$, and the weight for the $n$\textit{-th} scale layer is defined as
\begin{equation}
w(l, \alpha)=\frac{(1-\cos (\pi \operatorname{clamp}(\alpha-n, 0,1))}{2},
\end{equation}
where $k$ is current training iteration and ${s}$ is a hyper-parameter.
In this case, the high-scale weights are minimal at the beginning and are gradually risen.

% \begin{figure*}[!t]
%     \centering
%     \includegraphics[width=\textwidth]{fig/align.pdf}
%     \caption{Illustration of Transmission-guided Flow Rectification (TFR) module. In order to capture the time-frequency more accurately, the original estimated flow is rectified under the guidance of transmission maps for the effective alignment.
%     }
%     \vspace{-3mm}
%     \label{fig3}
% \end{figure*}

\vspace{-3mm}
\subsection{Transmission-guided Flow Rectification}

\begin{figure*}[!t]
    \centering
    \includegraphics[width=\textwidth]{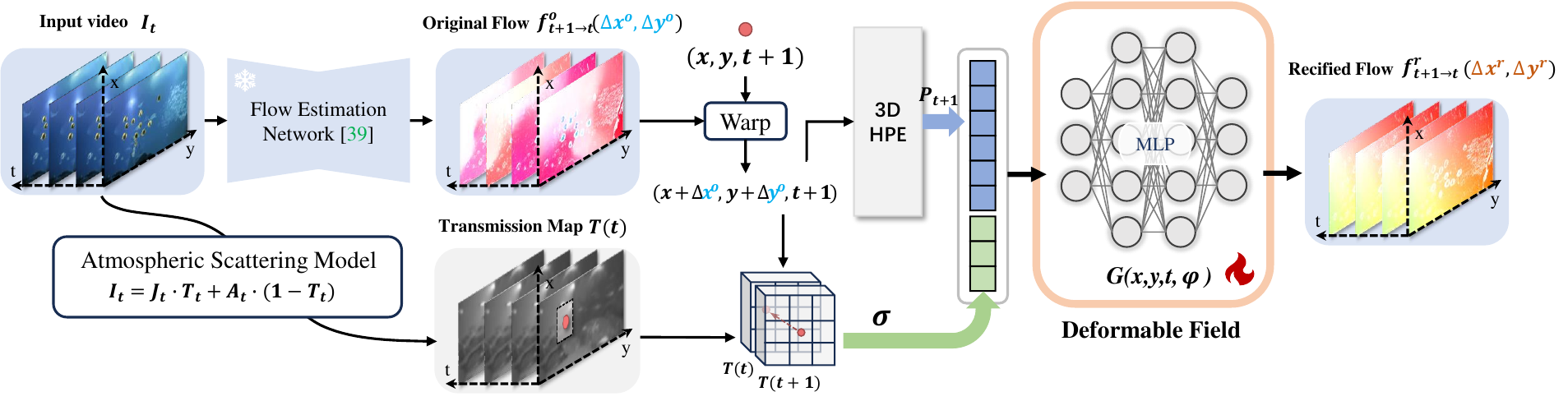}
    \caption{Illustration of Transmission-guided Flow Rectification (TFR) module. In order to capture the time-frequency more accurately, the original estimated flow is rectified under the guidance of transmission maps for the effective alignment.
    }
    \vspace{-3mm}
    \label{fig3}
\end{figure*}

To mitigate the motion impact on capturing temporal frequencies, it is natural to align temporal frames. However, existing optical flow estimations are mainly trained in open-air scenes, leading to that they are not so accurate in underwater scenarios. To address this issue, we introduce the Transmission-guided Flow Rectification (TFR) module, which corrects the directly estimated optical flow by the pretrained model. As shown in Figure~\ref{fig3}, considering two adjacent input frames $F(t)$ and $F(t+1)$, we first employ RAFT~\cite{Teed2020RAFTRA} to estimate the original optical flow $f_{t+1 \rightarrow t}^o(\Delta x^o, \Delta y^o)$. 

Then the initially warped grid is sent into the 3D multi-scale hash position encoding (HPE) to obtain the position embedding $P_{t+1}$, which is denoted as 
\begin{align}
P_{t+1} = HPE((x+\Delta x^o, y+\Delta y^o, t+1)).
\end{align}
Furthermore, we leverage the atmospheric scattering model, which is usually used in underwater modeling, to estimate the transmission maps $T_t$ and $T_{t+1}$ according to
\begin{equation}
   F_t = J_t \cdot T_t + A_t \cdot (1-T_t),
\end{equation}
where $J_t$ is the clear image and $A_x$ is the homogeneous background light. 

Since the medium transmission $T_t$ represents the proportion of scene radiance that reaches the camera after undergoing reflection within the medium, we serve the local feature of $T_t$ and $T_{t+1}$ as guidance, denoted as $\sigma$, and concatenate it with $P{t+1}$. Inspired by the deformable field~\cite{Brualla_2021}, given $\mathbb{C}(P_{t+1},\sigma)$, the rectified flow could be adaptively learned by MLP, which could be expressed by 
\begin{equation}
f^r_{t+1 \rightarrow t}(\Delta x^r, \Delta y^r) = MLP(\mathbb{C}(P_{t+1},\sigma)),
\end{equation}
where $\mathbb{C}$ is the concatenation operator.
In the end, we obtain the rectified flow $f_{t+1 \rightarrow t}^r(\Delta x^r, \Delta y^r)$ for more precise warp function.

\begin{figure*}[!t]
    \centering
    \includegraphics[width=\textwidth]{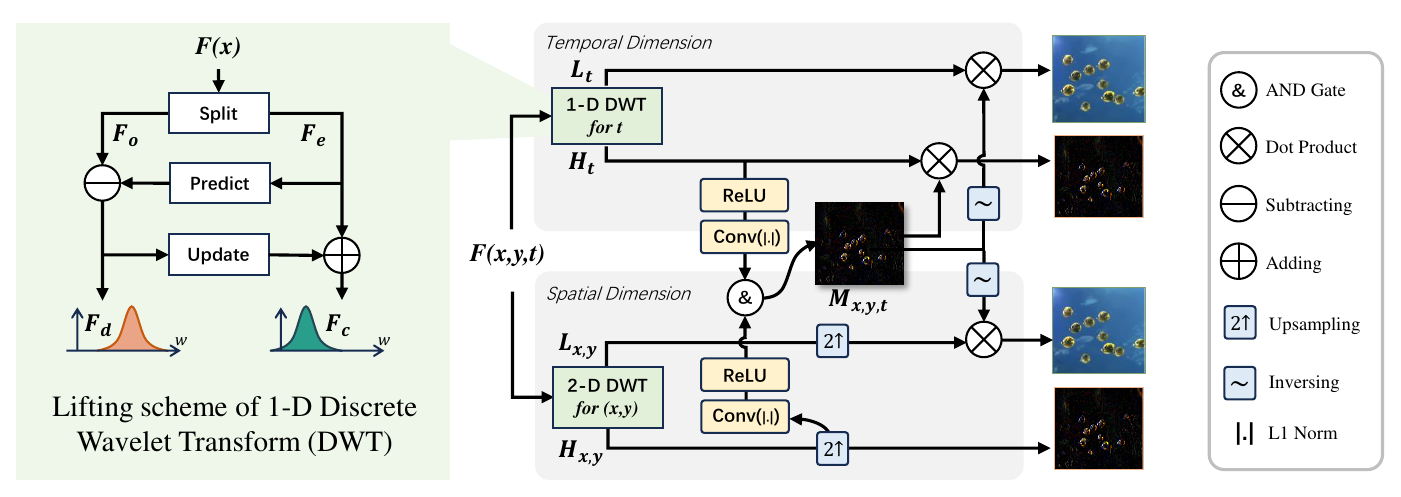}
    \caption{Overview of the Video Consistency-aware Wavelet (VC-Wave) block, which is wavelet-like transform for decoupling into basic contents, temporal inconsistent elements and motion details from video $F(x,y,t)$.
    }
    \vspace{-3mm}
    \label{fig4}
\end{figure*}

\vspace{-3mm}
\subsection{Video Consistency Learning}

In this section, we present the wave-based video continuity learning, which plays the core role in WaterWave. First, we introduce the structure of VC-Wave as a novel wavelet-like transform. Next, the training strategy and optimization objective in video consistency learning are also introduced. 

\noindent\textbf{Video Consistency-aware Wavelet.}
The Video Consistency-aware Wavelet (VC-Wave) implements a wavelet-like transform by introducing temporal consistency prior into the lifting scheme~\cite{Daubechies_Sweldens_2005}, which is a fast and in-place implementation of wavelet.
The lifting scheme comprises three fundamental steps: splitting, prediction, and updating. Taking one-dimensional case as example, as Figure~\ref{fig4} showing, we first split $F(x)$ into even and odd parts, $F_e$ and $F_o$. Then $F(x)$ is decoupled into high-frequency part $F_d$ and low-frequency ones $F_c$ by
\begin{equation}
F_d = F_o - Predict(F_e), \qquad F_c = F_e - Update(F_d),
\end{equation}
where $Predict()$ and $Update()$ denote prediction filter and update filter, respectively, and they are usually of simple linear forms.
The lifting scheme ensures decent reconstruction, and more importantly, it can be easily adapted to different wavelets by simply adjusting the parameters in the predict and update steps.
For example, in Haar wavelet, $Predict()$ and $Update()$ is set as $[0.5,0.5]$ and $[1,1]$. Other wavelet transforms have different forms referring~\cite{Daubechies_Sweldens_2005}.

Building on the lifting scheme and the temporal consistency prior in Section 2, we design the VC-Wave block. As shown in Figure~\ref{fig4}, given the video $F(x,y,t)$ fitted by the INR network, we apply the lifting scheme to perform 2-D DWT on the spatial domain and 1-D DWT on the temporal domain of $F(x,y,t)$ respectively, which can be represented as follows:
% \begin{equation}
% \centering
% L_{t}, H_{t} = DWT_{t}(F(x,y,t))), 
% \quad L_{x,y},H_{x,y} = DWT_{x,y}(F(x,y,t))), 
% \end{equation}
\begin{equation}
\scalebox{0.85}{$
L_t,H_t=\mathrm{DWT}_{t}\!\big(F(x,y,t)\big),\;
L_{x,y},H_{x,y}=\mathrm{DWT}_{x,y}\!\big(F(x,y,t)\big)
$}
\end{equation}
where $L_{t}, H_{t}, L_{x,y}, H_{x,y}$ are the low and high frequency of temporal and spatial dimension in the wavelet domain.
Subsequently, based on the temporal consistency prior, the local region with strong spatial high-frequency but weak temporal high-frequency amplitude often is taken as temporal inconsistency.
To extract the inconsistency component, denoted as $M_{x,y,t}$, we first choose the high-enough responses in time and the low-enough responses in space, which is performed through convolution layers and ReLU activation function. The process could be expressed as
% \begin{equation}
% M_{x,y,t} = \left\{ReLU\left(\lvert H_{t}\rvert - \beta_0\right)\right\} \& \left\{ReLU\left(-\lvert Conv(Up(H_{x,y}))\rvert + \beta_1\right)\right\},
% \end{equation}
\begin{equation}
\scalebox{0.83}{$
M_{x,y,t}
= \operatorname{ReLU}\!\bigl(\lvert H_t\rvert - \beta_0\bigr)
\;\&\;
\operatorname{ReLU}\!\bigl(-\,\lvert \mathrm{Conv}(\mathrm{Up}(H_{x,y}))\rvert + \beta_1\bigr)
$}
\end{equation}

where $\&$ is AND gate operation, $|.|$ denotes L1 norm, and $Up$ is $2x$ up-sampling. The threshold $\beta_0$ and $\beta_1$ are set as $0.001$ and $0.01$. 
Note that considering the alignment errors in the pixel level, a extra convolution layer $Conv$ with $ 3 \times 3 $ kernels is applied for aggregating neighboring information.

% \begin{figure*}[!t]
%     \centering
%     \includegraphics[width=\textwidth]{fig/wave.pdf}
%     \caption{Overview of the Video Consistency-aware Wavelet (VC-Wave) block, which is wavelet-like transform for decoupling into basic contents, temporal inconsistent elements and motion details from video $F(x,y,t)$.
%     }
%     \vspace{-3mm}
%     \label{fig4}
% \end{figure*}

\noindent\textbf{Optimization Object.}
% 在优化目标上，我们做两件事，第一件，约束输出图连续。第二件，让他在对E的重新拟合过程中，只学习内容和细节，而不学习其中的不连续性。
The optimization objective is developed for two principle, 1) guaranteeing the output video is temporally consistent, 2) only learning the basic content and motion details of UWIE enhanced video $V_t$ while not learning harmful temporal inconsistent elements of it.
To achieve the aforementioned objectives, we design three loss functions that jointly contribute during the training process.
For the first object, we apply VC-Wave to output video $F_t$, obtaining temporal inconsistent mask $M^F_{x,y,t}$. Then the temporal inconsistent elements could be extracted and regularized by following loss function  
\begin{equation}
Loss^{tc} = || M^F_{x,y,t} \cdot H^F_{t}  ||_1, 
\end{equation}
where $H^F_{t}$ denotes $H_{t}$ of $F_t$.
For the second object, we send enhanced video $V_t$ into the VC-Wave and get $M^V_{x,y,t}$. 
To relieve the temporal inconsistency from $V_t$, the network is driven by
% \begin{align}
% Loss^{detail} &= || H^F_{x,y} - H^V_{x,y} ||_1, \\
% Loss^{basic} &= \left( ||L^{F}_{t} - L^{V}_{t}||_1 + ||L^{F}_{x,y} - L^{V}_{x,y}||_1^1 \right)* M^V_{x,y,t}. 
% \end{align}
\begin{equation}
\scalebox{0.98}{$
\mathcal{L}^{\text{detail}} = \| H^F_{x,y} - H^V_{x,y} \|_{1}
$}
\end{equation}
\begin{equation}
\scalebox{0.98}{$
\mathcal{L}^{\text{basic}} =
\big( \|L^{F}_{t}-L^{V}_{t}\|_{1} + \|L^{F}_{x,y}-L^{V}_{x,y}\|_{1} \big)\, M^V_{x,y,t}
$}
\end{equation}

%% file: sec/5_exer.tex
\begin{table*}[t]
% \vspace{-4mm}
\caption{Quantitative comparison with other methods on DRUVA and UVOT400 datasets in metrics of CLIP-A~\cite{CLIPIQA} and NIMA~\cite{NIMA} on various UWIE method.}
\centering
\footnotesize
\resizebox{\linewidth}{!}{
\begin{tabular}{c|c|cccccccccc}
\toprule[1.2pt]
% \multicolumn{1}{c}{Datasets} & 
\multirow{2}{*}{\textbf{Datasets}} & 
\multirow{2}{*}{\textbf{Methods}} &
\multicolumn{2}{c}{\textbf{Boths-UIEB}} & 
\multicolumn{2}{c}{\textbf{Boths-UVE}} & 
\multicolumn{2}{c}{\textbf{DWN-UIEB}} & 
\multicolumn{2}{c}{\textbf{DWN-EUVP}} & 
\multicolumn{2}{c}{\textbf{FiveAPlus}} \\
% \midrule
% \multicolumn{1}{c}{-} & -& X & Y & X & Y & X & Y & X & Y & X & Y \\
& & CLIP-A & NIMA & CLIP-A & NIMA & CLIP-A & NIMA & CLIP-A & NIMA & CLIP-A & NIMA \\
\hline
\midrule
% \multirow{5}{*}{DRUVA} & Input & 0.2991 & 3.9402 &  0.2991 & 3.9402 &  0.2991 & 3.9402 & 0.2991 &3.9402&  0.2991 & 3.9402 \\
%  &UWIE &0.1874 & 4.5786& 0.2003 & 4.3912& 0.1487 & 4.9555 & 0.2829 & 4.2137 & 0.1937 &4.2539\\
%  &+CodeF &xxx & xxx & xxx & xxx & xxx &xxx& xxx & xxx & xxx &xxx \\
%  &+DVP &0.2114 & 4.5668 &0.2370 &4.4697 & 0.2082 & 4.8833 & 0.3841 & 4.0139 & ??? & ??? \\  %\rowcolor{gray!40} 
%  & \cellcolor{gray!40} +Ours &\cellcolor{gray!40}0.3177 &\cellcolor{gray!40}5.1813 & \cellcolor{gray!40}0.3146 & \cellcolor{gray!40}4.9323 & \cellcolor{gray!40}0.2995 & \cellcolor{gray!40} 5.3961&\cellcolor{gray!40} 0.3663 & \cellcolor{gray!40} 4.5363 & \cellcolor{gray!40}??? & \cellcolor{gray!40}??? \\
% \midrule
% \multirow{5}{*}{UVOT400}  & Input &0.3553 &5.2399& 0.3553 &5.2399 & 0.3553 &5.2399 & 0.3553 &5.2399 & 0.3553 &5.2399 \\
%  & UWIE &0.3048 & 5.1117 & 0.2810 & 5.2024 & 0.2394 &5.3360& 0.2628&5.3962 & 0.3555 & 5.1354 \\
% & +CodeF &xxx & xxx & xxx & xxx &xxx &xxx &xxx &xxx & xxx & xxx\\
%  & +DVP &0.2804 & 5.0776 & 0.2785 &5.0376 & 0.3059 & 4.9534 & 0.3300 &5.3449 & ??? & ??? \\   %\rowcolor{gray!40} 
% &\cellcolor{gray!40}  +Ours& \cellcolor{gray!40} 0.4111&\cellcolor{gray!40}5.3903 & \cellcolor{gray!40} 0.3865 & \cellcolor{gray!40}xxxx &\cellcolor{gray!40}0.3435 & \cellcolor{gray!40} 5.6224 & \cellcolor{gray!40} 0.3247 & \cellcolor{gray!40}5.4537 & \cellcolor{gray!40} xxxx& \cellcolor{gray!40} xxxx \\
\multirow{5}{*}{DRUVA} & Input &0.2991 &\textcolor{blue}{ 3.9402} &  0.2991 &\textcolor{blue}{ 3.9402} &  0.2991 &\textcolor{blue}{ 3.9402} & 0.2991 &3.9402&\textcolor{red}{  0.2991} & 3.9402 \\
 &UWIE &\textcolor{blue}{0.1874} & 4.5786&\textcolor{blue}{ 0.2003} & 4.3912&\textcolor{blue}{ 0.1487} & 4.9555 &\textcolor{blue}{ 0.2829} & 4.2137 &\textcolor{blue}{ 0.1937} &4.2539\\
 &+CodeF &0.2541 & 4.9336 & 0.2663 & \textcolor{red}{4.9462} & 0.2846 & 5.1326& \textcolor{red}{0.3844} &\textcolor{blue}{ 3.9128} & 0.2743 &4.3831\\
 &+DVP &0.2114 & 4.5668 &0.2370 &4.4697 & 0.2082 & 4.8833 & 0.3841 & 4.0139 & 0.2149 &  \textcolor{blue}{ 3.8936 }\\  %\rowcolor{gray!40} 
 & \cellcolor{gray!40} +Ours &\cellcolor{gray!40}\textcolor{red}{0.3177} &\cellcolor{gray!40}\textcolor{red}{5.1813} & \cellcolor{gray!40}\textcolor{red}{0.3146} & \cellcolor{gray!40}4.9323 & \cellcolor{gray!40}\textcolor{red}{0.2995} & \cellcolor{gray!40}\textcolor{red}{5.3961}&\cellcolor{gray!40} 0.3663 & \cellcolor{gray!40}\textcolor{red}{4.5363} & \cellcolor{gray!40}0.2943 & \cellcolor{gray!40}\textcolor{red}{4.8527} \\
\midrule
\multirow{5}{*}{UVOT400}  & Input &0.3553 &5.2399& 0.3553 &5.2399 &\textcolor{red}{ 0.3553} &5.2399 &\textcolor{red}{  0.3553} &\textcolor{blue}{5.2399} &0.3553&5.2399 \\
 & UWIE &0.3048 & 5.1117 & 0.2810 & 5.2024 &\textcolor{blue}{ 0.2394} &5.3360&\textcolor{blue}{ 0.2628}&5.3962 & 0.3555 & 5.1354 \\
& +CodeF &\textcolor{blue}{0.2490} &\textcolor{blue}{ 4.9417} &\textcolor{blue}{ 0.2646} & 5.2305 &0.2491 &5.5403 &0.2973 &5.2595 &\textcolor{blue}{ 0.2896} &\textcolor{red}{ 5.4739}\\
 & +DVP &0.2804 & 5.0776 & 0.2785 &\textcolor{blue}{5.0376} & 0.3059 &\textcolor{blue}{ 4.9534} & 0.3300 &5.3449 & 0.3161 &\textcolor{blue}{  5.0220} \\   %\rowcolor{gray!40} 
&\cellcolor{gray!40}  +Ours& \cellcolor{gray!40}\textcolor{red}{ 0.4111}&\cellcolor{gray!40}\textcolor{red}{5.3903} & \cellcolor{gray!40}\textcolor{red}{ 0.3865} & \cellcolor{gray!40}\textcolor{red}{5.5666} &\cellcolor{gray!40}0.3435 & \cellcolor{gray!40}\textcolor{red}{ 5.6224} & \cellcolor{gray!40} 0.3247 & \cellcolor{gray!40}\textcolor{red}{5.4537} & \cellcolor{gray!40}\textcolor{red}{ 0.4276}& \cellcolor{gray!40}5.4599 \\
\bottomrule[1.2pt]
\end{tabular}
}
\label{tab2}
\vspace{-2mm}
\end{table*}

\vspace{-4mm}
\begin{figure*}[t]
\centering
\includegraphics[width=\textwidth]{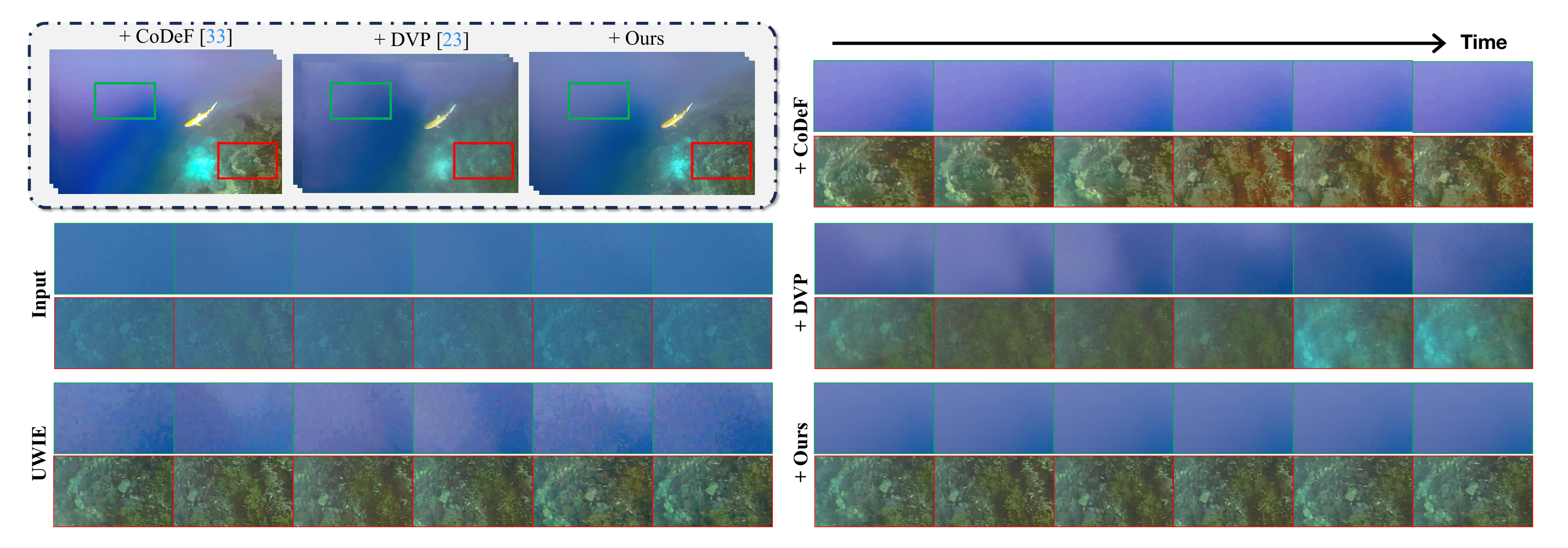}
\caption{Comparisons with other blind video continuity methods in the real underwater scene. Taking the input as the temporal continuity reference and UWIE frame-by-frame as the spatial enhancement reference, our restored underwater video not only maintains temporal consistency but also ensures that the enhanced details are not deteriorated.
}
\vspace{-3mm}
\label{fig5}
\end{figure*}

\vspace{5mm}
\section{Experiments}
% \vspace{-1mm}
\subsection{Experimental Setup}
\textbf{Datasets.} To evaluate the visual quality of restored video, two underwater video datasets, DRUVA~\cite{Varghese_2023_ICCV} and UVOT400~\cite{Alawode2023ImprovingUV} are adopted. The DRUVA~\cite{Varghese_2023_ICCV} dataset consists of 20 clips of video and we select 100 frame from each scene, where totally 2000 frame are contained. 
UVOT400~\cite{Alawode2023ImprovingUV} is for underwater object tracking task, which presents diverse marine animals and environment. We select 1000 frame consisting of ten scenes from UVOT400 for training.

\noindent\textbf{Implementation details.}
To demonstrate the effectiveness of our method, we use five UWIE models, Boths~\cite{boths_2023} pretrained on UIEB~\cite{Li_Guo_2019} and UVE-38K dataset~\cite{qi2021underwater}, DeepWaterNet (DWN)~\cite{sharma2021wavelengthbased} pretrained on UIEB~\cite{Li_Guo_2019} and EUVP dataset~\cite{Islam_Xia_Sattar_2020} and FiveAPlus~\cite{jiang2023five} as our benchmarks.
We adopt Adam optimizer and the initial learning rates is set to $1 \times 10^{-3}$ and halved at 15,000 iteration.

\begin{figure*}[!t] 
\centering
\begin{minipage}[t]{0.57\textwidth}
\captionof{table}{Performance comparisons of different trackers equipped by UWIE and our method.}
\label{tab:trackers}
\footnotesize
\resizebox{\linewidth}{!}{
\begin{tabular}{c|c|c|ccc}
\toprule[1.2pt]
\textbf{Method} & \textbf{Venue} & \textbf{Config}& \textbf{Success} & \textbf{Norm Pre.} & \textbf{Precision} \\
\midrule
\multirow{3}{*}{\thead{ARTrack\\~\cite{Wei_2023_CVPR}}} &\multirow{3}{*}{\thead{CVPR\\2023}} & Base & \textcolor{red}{90.1} &  \textcolor{blue}{97.6} &  \textcolor{red}{90.0} \\
 & & +UWIE & \textcolor{blue}{89.4(-0.7)} & 97.7(+0.1) & \textcolor{blue}{88.1(-1.9)}\\
 & & \cellcolor{gray!40}+Ours &\cellcolor{gray!40}89.5(-0.6) & \cellcolor{gray!40}\textcolor{red}{98.1(+0.5)} &\cellcolor{gray!40}88.1(-1.9) \\
\midrule
\multirow{3}{*}{\thead{DropTrack\\~\cite{dropmae2023}}} &\multirow{3}{*}{\thead{CVPR\\2023}} & Base & \textcolor{blue}{72.7} &\textcolor{blue}{78.3} & \textcolor{blue}{69.0} \\
 & & +UWIE &  \textcolor{red}{82.3(+9.6)} &  \textcolor{red}{89.1(+10.8)} &  \textcolor{red}{82.6(+13.6)}\\
 & & \cellcolor{gray!40}+Ours &\cellcolor{gray!40}75.1(+2.4) &\cellcolor{gray!40}82.7(+4.4) & \cellcolor{gray!40}73.1(+4.1) \\
\midrule
\multirow{3}{*}{\thead{AiATrack\\~\cite{gao2022aiatrack}}} &\multirow{3}{*}{\thead{ECCV\\2022}} & Base &  \textcolor{blue}{67.1} &  \textcolor{red}{74.3} & 56.0 \\
 & & +UWIE & 67.5(+0.4) &  \textcolor{blue}{72.6(-1.7)} &  \textcolor{blue}{55.0(-1.0)}\\
 & & \cellcolor{gray!40}+Ours &\cellcolor{gray!40}\textcolor{red}{68.6(+1.5)} &\cellcolor{gray!40}72.7(-1.6) &\cellcolor{gray!40}\textcolor{red}{59.3(+3.3)} \\
\midrule
\multirow{3}{*}{\thead{MAT\\~\cite{MAT}}} &\multirow{3}{*}{\thead{CVPR\\2023}} & Base & 74.0 & 90.1 & 61.0 \\
 & & +UWIE & \textcolor{blue}{67.0(-7.0)} & \textcolor{blue}{78.9(-11.2)} & \textcolor{blue}{59.9(-1.1)}\\
 & &\cellcolor{gray!40} +Ours &\cellcolor{gray!40}\textcolor{red}{ 77.1(+3.1)} & \cellcolor{gray!40}\textcolor{red}{90.9(+0.8)} &\cellcolor{gray!40}\textcolor{red}{70.7(+9.7)} \\
\midrule
\multirow{3}{*}{\thead{GRM\\~\cite{gao2023generalized}}} &\multirow{3}{*}{\thead{CVPR\\2023}} & Base & \textcolor{red}{72.8} & \textcolor{red}{81.7} &  \textcolor{blue}{67.1}\\
 & & +UWIE & 71.9(-0.9) & 80.9(-0.8) & 69.0(-1.9)\\
 & & \cellcolor{gray!40}+Ours &\cellcolor{gray!40}\textcolor{blue}{ 71.8(-1.0) }& \cellcolor{gray!40}\textcolor{blue}{77.6(-4.1)} & \cellcolor{gray!40}\textcolor{red}{72.0(+4.9)} \\
\midrule
\multirow{3}{*}{\thead{UOSTrack\\~\cite{UOSTrack}}} &\multirow{3}{*}{\thead{Ocean Eng\\2023}} & Base &  \textcolor{blue}{69.0} & \textcolor{blue}{ 77.0} &  \textcolor{blue}{60.6} \\
 & & +UWIE & 71.6(+2.6) & 82.0(+5.0) & 66.0(+5.4)\\
 & & \cellcolor{gray!40}+Ours & \cellcolor{gray!40}\textcolor{red}{83.7(+14.7)} & \cellcolor{gray!40}\textcolor{red}{93.0(+16.0)} &\cellcolor{gray!40}\textcolor{red}{80.3(+19.7) }\\ 
\bottomrule[1.2pt]
\end{tabular}
}
\end{minipage}
\hfill
%---------------- 右侧：图片 ----------------%
\begin{minipage}[t]{0.39\textwidth}
\centering
\vspace{2mm}
\includegraphics[width=\linewidth]{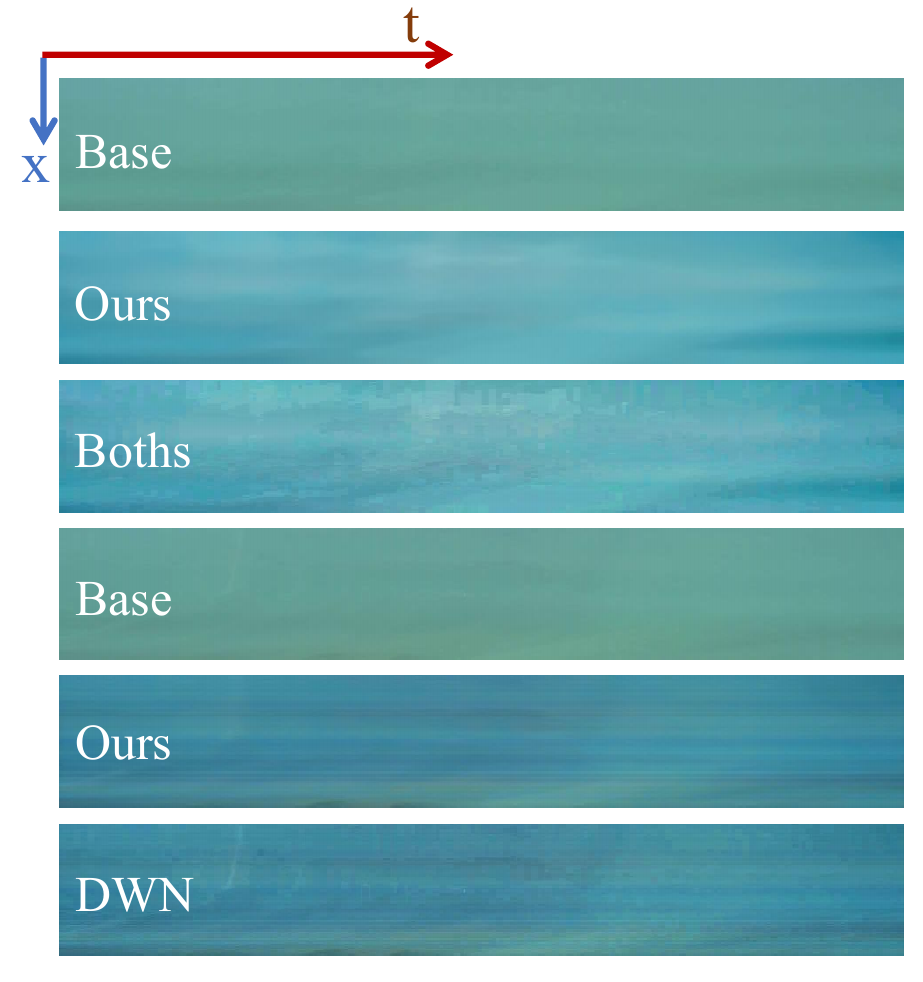} % 确保这个路径与文件存在
\captionof{figure}{Comparison of temporal profile. The temporal consistency is restored in different UWIEs.}
\label{fig6}
\end{minipage}
\end{figure*}

\begin{figure*}[t]
\centering
\includegraphics[width=\textwidth]{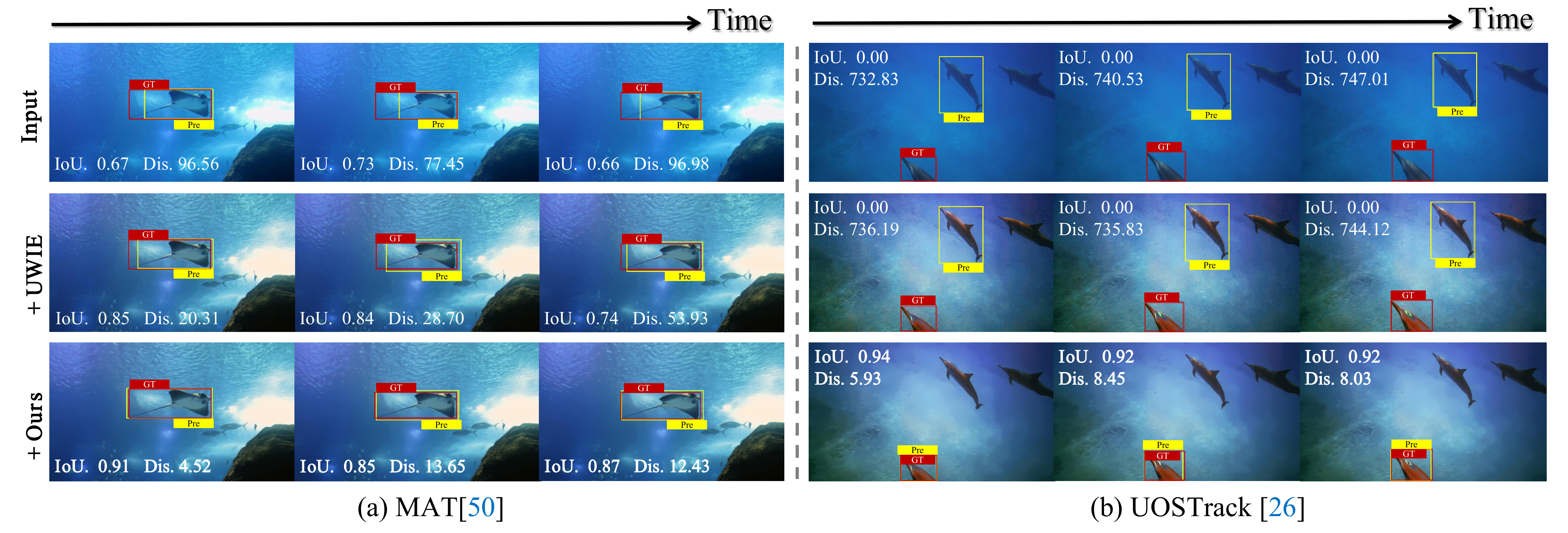}
\caption{
Visualization comparisons of our method on underwater video tracking tasks demonstrates more accurate tracking results. In (b), due to the enhancement of temporal consistency, we successfully find the truth target that are originally lost.
}
\vspace{-3mm}
\label{fig77}
\end{figure*}

\vspace{-3mm}
\subsection{Performance Comparison}
\noindent\textbf{Effectiveness of human visual perception.}
To demonstrate the effectiveness of our experiments, we conduct quantitative and qualitative evaluations. Because obtaining underwater pairwise video is hard, we adopt no-reference quality metrics, CLIP-A~\cite{CLIPIQA} and NIMA~\cite{NIMA}. As shown in Table~\ref{tab2}, our method usually outperforms previous methods in terms of subjective quality. Furthermore, we conduct visual comparisons in real underwater scenarios, as shown in Figure~\ref{fig5}. It can be observed that while CoDeF maintains good temporal continuity, it suffers from spatial distortion. DVP enhances continuity but reintroduces degradation information. In the contrary, our method preserves the original enhancement in the spatial domain and maintains temporal continuity similar to the original video. Moreover, we examine a row and track changes over time for comparing from temporal profile. As shown in Figure~\ref{fig6}, the profile of UWIE occurs flickering artifacts and our method mitigates this problem a lot.

\noindent\textbf{Effectiveness of machine visual perception.}
To further demonstrate the effectiveness of our method in videos, we compare our method with UWIE in video object tracking tracking, as illustrated in Figure~\ref{fig77} and Table \textcolor{red}{3} Interestingly, we find that in some trackers, such as MAT, AiATrack, and ARTrack,  solely UWIE results can lead to performance drops due to the lack of temporal consistency. 
Fortunately, our method, via compensating for the temporal consistency, successfully aids in downstream video tasks. This observation may offer meaningful insights for the video low-for-high research in the future.

\vspace{-4mm}
\subsection{Ablation Study and Discussion}
\textbf{Effectiveness of Transmission Guided Flow Rectification.}
Due to light scattering in underwater environments, conventional optical flow estimation methods often struggle to achieve accurate motion alignment. Our designed flow rectification module incorporates the underwater transmission map as guidance, which improves the capture of underwater objects. As shown in Figure~\ref{fig9}, the optical flow is more accurate when combined with the transmission map, resulting in smaller alignment errors, particularly around swimming fish.
Furthermore, it can be observed that the estimated optical flow between two neighboring frames are distinct, which may struggle to maintain temporal consistency underwater. Our optical flow, on the other hand, exhibits better temporal coherence and have potential for restoring consistent video.

% \begin{figure*}[t]
% \centering
% \includegraphics[width=\textwidth]{fig/vot.pdf}
% \caption{
% Visualization comparisons of our method on underwater video tracking tasks demonstrates more accurate tracking results. In (b), due to the enhancement of temporal consistency, we successfully find the truth target that are originally lost.
% }
% \vspace{-3mm}
% \label{fig77}
% \end{figure*}

\begin{figure*}[t]
\centering
\includegraphics[width=\textwidth]{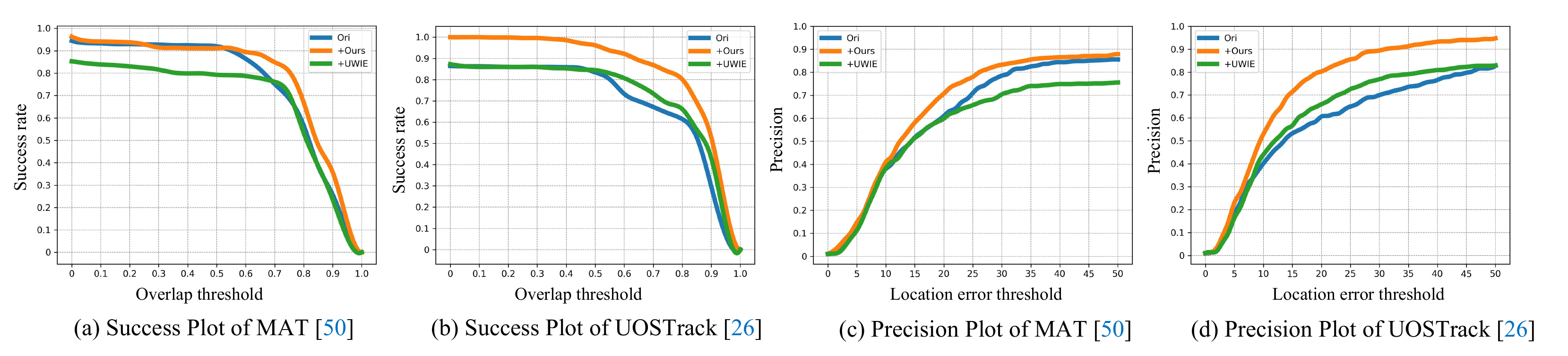}
\caption{Evaluations comparisons of different trackers using UVOT400 dataset. 
Our method brings significant performance improvements.
}
\vspace{-5mm}
\label{fig6b}
\end{figure*}

\begin{figure*}[t]
\centering
\includegraphics[width=\textwidth]{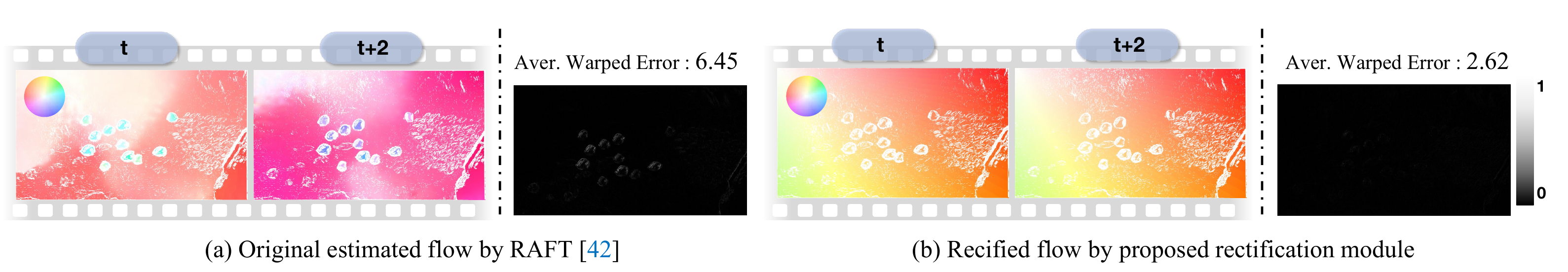}
\caption{
Comparison of estimated flow~\cite{Teed2020RAFTRA} and our rectified flow. Our rectified flow have better temporal coherence for more faithful reconstruction of continuous videos.
}
\vspace{-2mm}
\label{fig9}
\end{figure*}

% \begin{figure}[t]
% \centering
% \includegraphics[width=\textwidth]{eccv_fig/mask.pdf}
% \vspace{-4mm}
% \caption{
% The visualization of the temporal inconsistency. The temporal high frequency (e) includes temporal inconsistency and spatial details, and the spatial high frequency (d)
% mainly includes spatial details. Through the VC-Wave, the temporal inconsistency are represented in (f). After regularizing (f), the output (c) recovers decent temporal consistency (see red boxes) and retains motion details (see fish).
% }
% \vspace{-2mm}
% \label{fig10}
% \end{figure}

\begin{figure*}[!t]
\centering
\includegraphics[width=\textwidth]{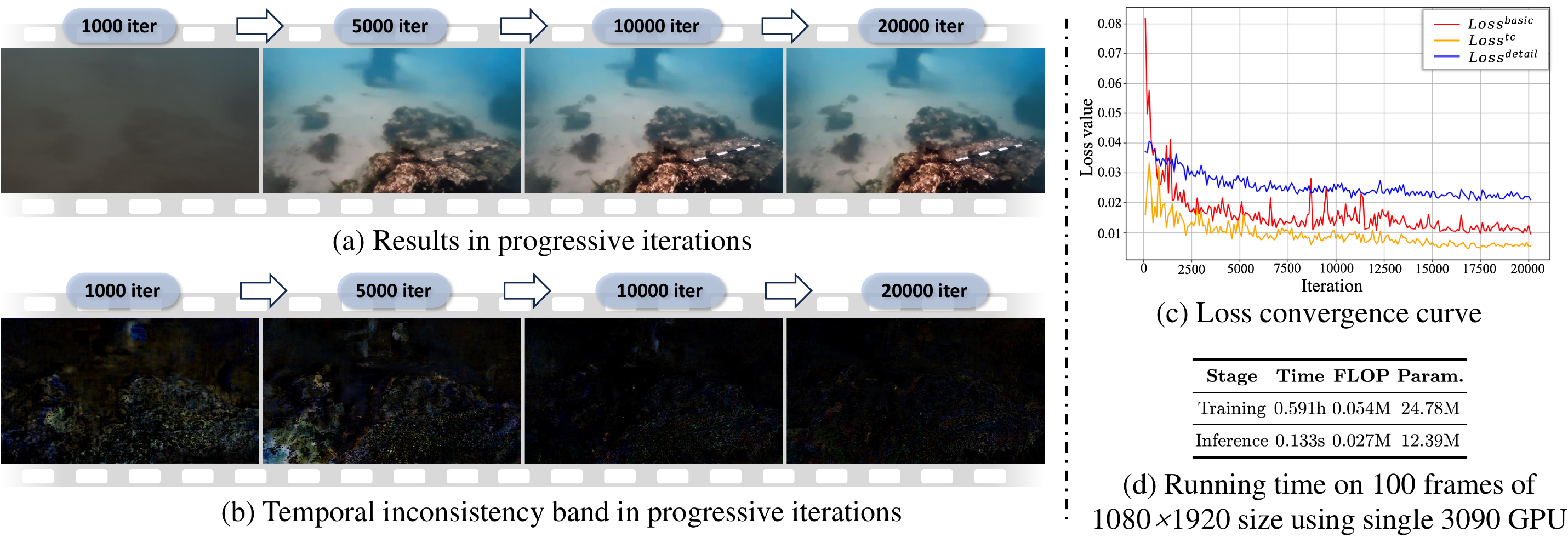}
\vspace{-3mm}
\caption{(a) represents spatial fidelity, while (b) represents temporal continuity. It can be observed that the fitting of temporal continuity lags behind the fitting of spatial information. (c) and (d) display loss convergence and efficiency of our network.
}
\vspace{-2mm}
\label{fig11}
\end{figure*}

% \noindent\textbf{Temporal inconsistency visualization.} 
% In this work, based on the temporal consistency prior, we design the VC-Wave block, which could decouple temporal inconsistency as mask, $M_{x,y,t}$. To valid its effectiveness, we visualize the mask as shown in Figure~\ref{fig10},

\noindent\textbf{Analysis of progressive iterations.}
In this work, we employ the INR as the primary framework for function fitting. To better understand the learning process of the network, we visualize the results and the temporal high-frequency components during progressive iterations in Figure~\ref{fig11}. It can be observed that at the 5000-th iteration, spatial information is well fitted, but the temporal dimension is not consistent, where many temporal high-frequencies occur. As following iterations, the temporal high-frequency is gradually regularized, bringing temporal consistency. In other words, the network initially focuses on fitting spatial information and then gradually addresses the consistency of time.

\section{Limitations} \label{seclimit}
\vspace{-1mm}
Although our method allows for customized temporal recovery for each scenario by fitting a set of specific parameters, increasing the number of scenes unavoidably prolongs training time. Unlike existing image restoration networks that share a common set of parameters across multiple scenes, thus not requiring additional training time, our method combines the characteristics of implicit representation networks and image processing networks.
To address the challenge of lacking video labels for training video restoration networks, a simple solution is proposed: our method can provide pseudo-video ground truth for video restoration networks. 
This approach partially resolves the issue of training video restoration networks due to the absence of video labels. 
In future, we will tap this potential of generating pseudo-labels to help train video models when lacking of paired data, which has great significance in the practical application.

\section{Conclusion}
In this work, we propose WaterWave for the first learning-based framework in enhancing underwater videos, notably improving their temporal consistency. This innovative approach addresses the long-standing challenge in deep learning-based underwater video enhancement, mainly stemming from the scarcity of underwater video pairs. 
In WaterWave, VC-Wave block and TFR module are proposed for decoupling and effectively eliminating temporally discontinuous elements in videos. Significant experiments are performed to show temporal consistency improvements in underwater image enhancements.
Furthermore, we explore the performance of our method on downstream underwater video tasks and find that it significantly enhances the accuracy of video tracking, which provides valuable exploration for video low-for-high tasks.